\documentclass[]{bytedance_seed}

\usepackage{wrapfig}
\usepackage{booktabs}
\usepackage{amsfonts}
\usepackage{tabularx}
\usepackage{array}
\usepackage[toc,page,header]{appendix}

\usepackage{minitoc}

\title{ One Model, Multiple Compute Budgets: Multi-Ratio ConceptMoE for Flexible Compute Across Turns  }

\author[1,2,*]{Yiyang Fu}
\author[2]{Zihao Huang}
\author[2,3,*]{Dongchen Han}
\author[1]{Shukai Gong}
\author[4]{Zhen Dong}
\author[2,\dagger]{Qiyang Min}
\author[1,\dagger]{Daquan Zhou}

\affiliation[1]{Peking University}
\affiliation[2]{ByteDance Seed}
\affiliation[3]{Tsinghua University}
\affiliation[4]{University of California—Santa Barbara}

\contribution[*]{Work done at ByteDance Seed}
\contribution[\dagger]{Corresponding authors}

\abstract{
Large language models typically support different inference budgets through
separate model sizes or fixed compute configurations.
In multi-turn interactions, however, compute demand can vary substantially
across turns.
Using a high-compute model for every turn wastes computation, while routing
among independent models may require rebuilding model-specific history states
after switches.
We therefore ask whether a single shared model can support multiple compute
budgets that can be flexibly assigned across turns.
We introduce \textbf{Multi-Ratio ConceptMoE}, which jointly supports
token-to-concept compression ratios of 2, 4, and 8 (CR2/CR4/CR8) within a
shared backbone.
Lightweight full-token modules reduce compression-independent cost,
ratio-specific boundary routers support distinct compression levels, and
cross-ratio training enables mixed-ratio processing within the same context.
Across PTBench and SFTBench, the three fixed modes outperform their
compute-matched MoE baselines in all \textbf{6} comparisons while remaining
competitive with independently trained fixed-ratio specialists.
On CodeChat and CoSQL, mixed-ratio inference improves the
accuracy--compute trade-off, outperforming the Large MoE by
\textbf{5.87} and \textbf{2.68} percentage points while reducing FLOPs by
\textbf{41.1\%} and \textbf{12.3\%}, respectively.
Under the same turn-wise budget schedule, Multi-Ratio ConceptMoE further
reduces FLOPs by \textbf{78.4\%} and \textbf{46.9\%} relative to routing
among independent models.
These results demonstrate that a single shared model can provide multiple
compute budgets for flexible compute allocation across turns.
}

\date{\today}
\correspondence{Qiyang Min at \email{minqiyang@bytedance.com}, Daquan Zhou at \email{zhoudaquan21@gmail.com}}

\begin{document}
\maketitle


\section{Introduction}

Large language models exhibit a natural trade-off between prediction quality and inference cost~\cite{chen2023frugalgpt}.
 In real-world multi-turn interactions, however, the computational workload is intrinsically non-uniform across conversational turns. An initial request often demands intense computation for complex reasoning or code generation, whereas subsequent turns frequently involve lightweight edits, local refinements, or follow-up clarifications~\cite{schuster2022confident, raposo2024mixture}. This dynamic nature of human-AI interaction calls for a paradigm shift from static inference to turn-wise adaptive compute allocation.
 


\begin{figure}
    \centering
    \includegraphics[width=1.0\linewidth]{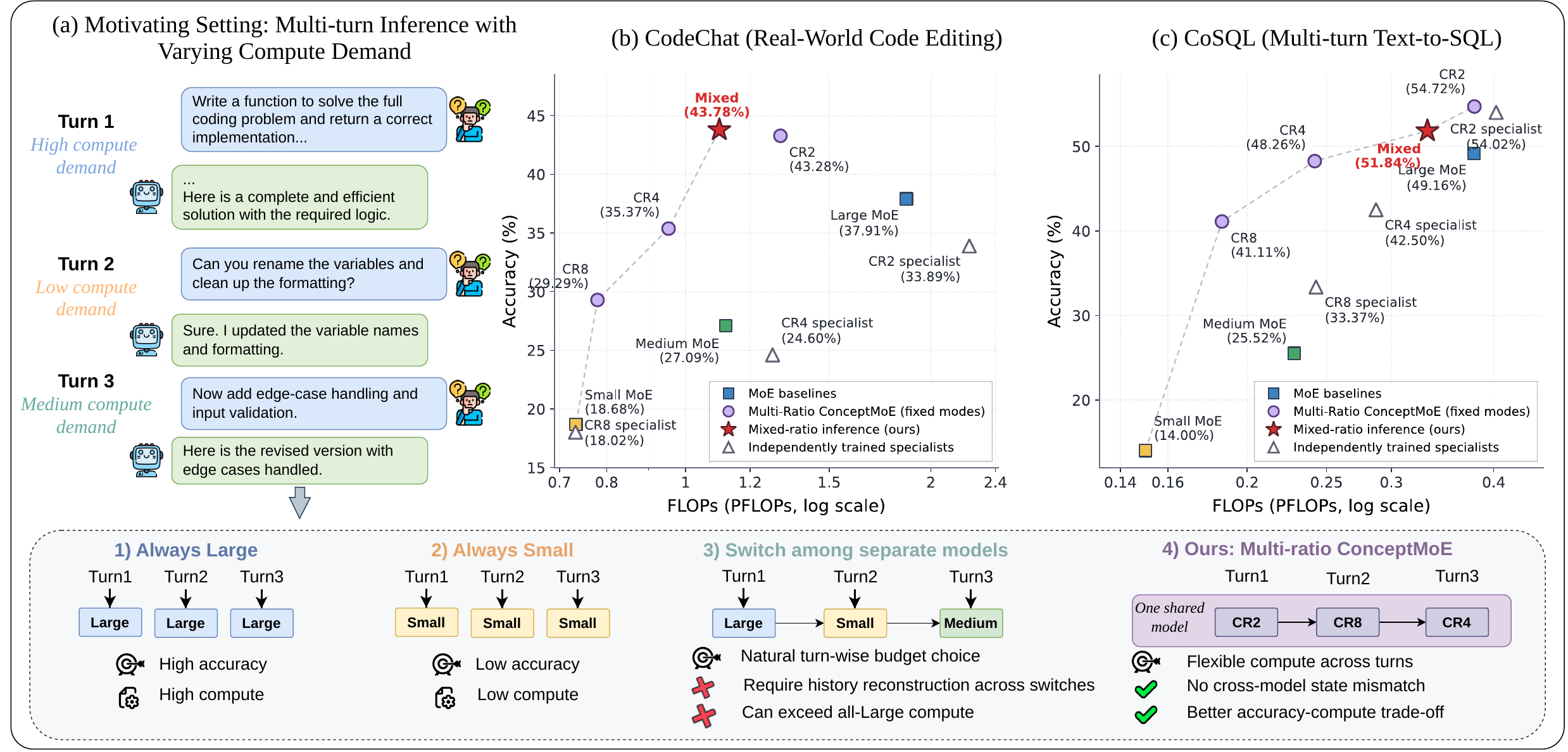}
    \caption{
    \textbf{Motivation and accuracy--compute trade-offs of Multi-Ratio ConceptMoE.}
    \textbf{(a)} Multi-turn interactions naturally exhibit varying compute
    demands across turns.
    Using a fixed Large or Small model leads to an unfavorable
    accuracy--compute trade-off, while switching among separately deployed
    models may require rebuilding model-specific history states.
    Multi-Ratio ConceptMoE instead supports different compute budgets within
    a single shared model.
    \textbf{(b,c)} On CodeChat and CoSQL, mixed-ratio inference lies on the
accuracy--compute Pareto frontier of the compared fixed-ratio ConceptMoE
modes, independently trained fixed-ratio specialists, and MoE baselines.
    }
    \label{fig:teaser}
\end{figure}

A natural strategy is to route requests among separately deployed Large,
Medium, and Small models~\cite{ong2025routellm}.
However, independent models maintain model-specific key-value (KV) states that
cannot be directly reused across models.
Switching models during a stateful interaction can therefore require rebuilding
the target model's history state from the preceding context
(Fig.~\ref{fig:teaser}(a)).
As the dialogue context grows, this additional history processing can offset
part of the compute saved by routing to a smaller model.
This motivates a compelling research question: 
\textit{Can we build a single shared model that supports multiple compute budgets that can be flexibly assigned across turns?}

In this paper, we answer this question with token-to-concept compression, providing a natural basis for exposing different compute budgets within a single model.
Recent concept models such as DLCM~\cite{DBLP:journals/corr/abs-2512-24617} and ConceptMoE~\cite{DBLP:journals/corr/abs-2601-21420} compress consecutive token
representations into a shorter sequence of semantic concepts, allowing the
main backbone to process fewer representations.
In ConceptMoE, the target compression ratio directly controls the overall compute budget by changing the number of concept representations processed by the backbone.
Existing fixed-ratio models, however, support only one such budget after training. We extend this setting to multiple compression-based budgets that can be used
across interaction turns.

We introduce \textbf{Multi-Ratio ConceptMoE}, which supports multiple compute budgets
within a single shared model.
The design addresses two challenges.
First, the token-level encoder and decoder operate at full token resolution, so
their computation does not decrease proportionally as the compression ratio
increases and can create a substantial compression-independent compute floor.
We therefore make the full-token modules lightweight and concentrate more model
capacity in the compression-scaled concept backbone, widening the effective
compute range across compression ratios.
Second, strong performance at each individual compression ratio does not
necessarily translate to effective mixed-ratio inference.
Multi-Ratio ConceptMoE uses lightweight ratio-specific boundary routers to
jointly support CR2, CR4, and CR8 within the shared backbone.
We further introduce \textbf{cross-ratio training}, which applies different compression ratios within the same training sequence
so that the shared model learns to handle mixed-ratio representations.
Together, these designs allow one checkpoint to serve as multiple fixed
operating points and to perform mixed-ratio inference across turns.

Our experiments show that Multi-Ratio ConceptMoE provides both strong fixed
operating points and improved accuracy--compute trade-offs.
Across PTBench and SFTBench evaluation suites
(see Appendix~\ref{app:general_benchmarks} for benchmark composition),
its CR2, CR4, and CR8 modes outperform the corresponding compute-matched
Large, Medium, and Small MoE baselines in all 6 comparisons, while remaining
competitive with independently trained fixed-ratio specialists.

More importantly, mixed-ratio inference improves the accuracy--compute
trade-off on both multi-turn tasks (Fig.~\ref{fig:teaser}(b,c)).
Compared with the Large MoE, it improves accuracy by \textbf{5.87} and
\textbf{2.68} percentage points while reducing FLOPs by \textbf{41.1\%} and
\textbf{12.3\%} on CodeChat and CoSQL, respectively; on CodeChat, it also
strictly dominates the fixed CR2 mode.
Furthermore, under the same turn-wise compute-budget schedule, the shared
Multi-Ratio model reduces FLOPs by \textbf{78.4\%} on CodeChat and
\textbf{46.9\%} on CoSQL compared with routing among independent models,
largely by avoiding repeated history re-prefill after model switches.

\paragraph{Contributions.}
Our main contributions are summarized as follows:
\begin{itemize}
    \item \textbf{Single-model flexible compute for multi-turn inference.}
    We study a shared-model approach to turn-wise compute allocation, where a single model supports multiple compute budgets that can be flexibly used across turns.

    \item \textbf{Multi-Ratio ConceptMoE.}
We extend token-to-concept compression to multiple compute budgets within a
shared model, combining lightweight full-token processing, ratio-specific
boundary routers, and cross-ratio training to support mixed-ratio inference.

    \item \textbf{Improved accuracy--compute trade-offs.}
    Multi-Ratio ConceptMoE provides strong fixed operating points across compute
budgets, while mixed-ratio inference further improves the accuracy--compute
trade-off on CodeChat and CoSQL.
    Under the same budget schedule, the shared model also avoids substantial
    history re-prefill overhead incurred by routing among independent models.
\end{itemize}

\section{Related Work}

\subsection{Adaptive and Multi-Budget Computation}

Prior work has explored several ways to support different inference budgets,
including training models of different sizes, model-level routing or
cascading~\cite{chen2023frugalgpt, ong2025routellm}, and varying computation
within a single model through early exiting, dynamic depth, elastic subnetworks,
or adaptive MoE activation~\cite{elbayad2019depth, raposo2024mixture,
devvrit2024matformer, cai2024flextron, DBLP:conf/emnlp/ZengMG0D24, DBLP:journals/corr/abs-2509-21892}.
Recent budget-driven methods such as BUDDY~\cite{zhou2026buddybudgetdrivendynamicdepth} further
support multiple user-specified compute budgets within a single model by
dynamically varying the executed layers.
These approaches expose different accuracy--compute operating points by varying
model depth, active subnetworks, or activated parameters.
Multi-Ratio ConceptMoE controls computation through representation
granularity:
the shared model changes how many concept representations are processed by the
backbone through the token-to-concept compression ratio.
We further study how these compute budgets can be used within the same context
and across interaction turns.

\subsection{Latent Concept Models and Token-to-Concept Compression}

The most closely related work focuses on latent concept models.
\textbf{Dynamic Large Concept Models (DLCM)} learns latent semantic boundaries
and compresses variable-length token spans into concepts, shifting more
computation from token space to a compact concept space~\cite{DBLP:journals/corr/abs-2512-24617}.
Its compression-aware scaling analysis further studies how computation can be
allocated among compression ratio, token-level capacity, and concept-level
reasoning capacity.

Subsequently, \textbf{ConceptMoE}~\cite{DBLP:journals/corr/abs-2601-21420} introduces a learnable chunking module that
predicts content-adaptive concept boundaries under a target compression ratio.
ConceptMoE further combines token-to-concept compression with a
Mixture-of-Experts backbone to control total parameters and activated FLOPs,
enabling a more direct comparison between concept-level and conventional
token-level computation.

Our work builds directly on this token-to-concept compression paradigm and
further considers the joint support and composition of multiple compression
budgets.
Existing concept models typically operate under a predefined overall
compression budget, whereas Multi-Ratio ConceptMoE exposes multiple
compression-based compute operating points within a single shared model.
We jointly train multiple target compression ratios and use cross-ratio
training to expose the shared backbone to mixed-granularity concept
representations.
In this way, concept models are extended to support multiple compute operating
points, allowing us to study how these modes can be composed within the same
context and across interaction turns, as well as the resulting
accuracy--compute trade-offs.

\subsection{Efficient Multi-Turn Inference}

An important challenge in multi-turn inference is efficiently processing an
ever-growing interaction history.
Prior work has addressed this problem through context compression, KV-cache
compression and quantization, cache selection or eviction, and more efficient
long-sequence attention~\cite{rae2019compressive, beltagy2020longformer, zhang2023h2o, li2024snapkv, liu2024kivi, kwon2023efficient}.
These approaches mainly reduce the cost of representing, storing, or attending
to historical context; complementary inference-time techniques have also
reduced autoregressive generation cost, e.g., through speculative
decoding~\cite{leviathan2023fastinferencetransformersspeculative,gong2025tppsdacceleratingtransformerpoint}.
Another line of work adapts inference compute through model routing or
cascading, selecting among models of different capacities according to the
input or task~\cite{chen2023frugalgpt, ong2025routellm}.
Such approaches provide multiple compute levels through separate model
instances.
In stateful multi-turn serving, however, cached representations are
model-specific, so switching between models may require reconstructing the
history state for the newly selected model.

Our work complements these directions by realizing multiple
compression-based compute budgets within a single shared model, allowing
different budgets to be used across turns without switching between separate
model instances.

\section{Approach}
\label{sec:approach}

Our design extends ConceptMoE along two dimensions: widening the effective
compute range across compression ratios and enabling mixed-ratio processing.
The first requires reducing computation that remains at full token resolution,
while the second requires the shared backbone to handle representations produced
under different compression ratios.
We address these issues with lightweight full-token modules, ratio-specific
boundary routing, and cross-ratio training.
Figure~\ref{fig:method} provides an overview of the resulting architecture and
training scheme.

\subsection{Preliminaries: Token-to-Concept Compression}
\label{sec:preliminaries}

We briefly review the token-to-concept compression mechanism of
ConceptMoE~\cite{DBLP:journals/corr/abs-2601-21420}.
Given a sequence of $N$ token representations, a token-level encoder $\mathcal{E}$
produces contextualized representations while preserving the sequence length:
\begin{equation}
X=[x_1,\ldots,x_N]\in\mathbb{R}^{N\times d_e},
\qquad
H=\mathcal{E}(X)=[h_1,\ldots,h_N]\in\mathbb{R}^{N\times d_e}.
\label{eq:conceptmoe_encoder}
\end{equation}

The Chunk module then predicts content-adaptive concept boundaries from $H$.
Let $b_i\in\{0,1\}$ indicate whether position $i$ starts a new concept.
These boundary decisions partition the original sequence into $M$ contiguous
token chunks, denoted by $S_1,\ldots,S_M$.
Each chunk is compressed into one concept representation, producing
\begin{equation}
C=[c_1,\ldots,c_M]\in\mathbb{R}^{M\times d_c},
\qquad M\leq N.
\label{eq:concept_sequence}
\end{equation}
The concept backbone $\mathcal C$ processes the compressed sequence,
producing $\hat C=\mathcal C(C)\in\mathbb{R}^{M\times d_c}$.
DeChunk then restores token resolution, and the token-level decoder
$\mathcal D$ processes the resulting length-$N$ representation.
The overall computation is summarized as
\begin{equation}
X
\xrightarrow{\mathcal E} H
\xrightarrow{\mathrm{Chunk}} C
\xrightarrow{\mathcal C} \hat C
\xrightarrow{\mathrm{DeChunk}} Z
\xrightarrow{\mathcal D} \hat Z .
\label{eq:conceptmoe_pipeline}
\end{equation}

ConceptMoE controls the average compression level through a target ratio $r$,
encouraging $M\approx N/r$.
Since chunk boundaries are content-adaptive, the realized ratio
$\tilde r=N/M$ varies across inputs.
Ignoring the relatively small Chunk/DeChunk overhead and attention-map
computation, the total computation can be approximated as
\begin{equation}
F(r)\approx F_{\mathrm{full}}
+\frac{F_{\mathrm{concept}}}{r},
\label{eq:compute_decomposition}
\end{equation}
where $F_{\mathrm{full}}$ denotes full-token computation in the encoder and
decoder, and $F_{\mathrm{concept}}$ denotes the uncompressed computation of
the concept backbone.
As $r$ increases, concept-level computation decreases approximately as $1/r$,
while $F_{\mathrm{full}}$ remains unchanged.

\subsection{Multi-Ratio ConceptMoE Architecture}
\label{sec:multi_ratio_arch}

\begin{wrapfigure}[23]{r}{0.55\linewidth}
    \centering
    \includegraphics[width=\linewidth]{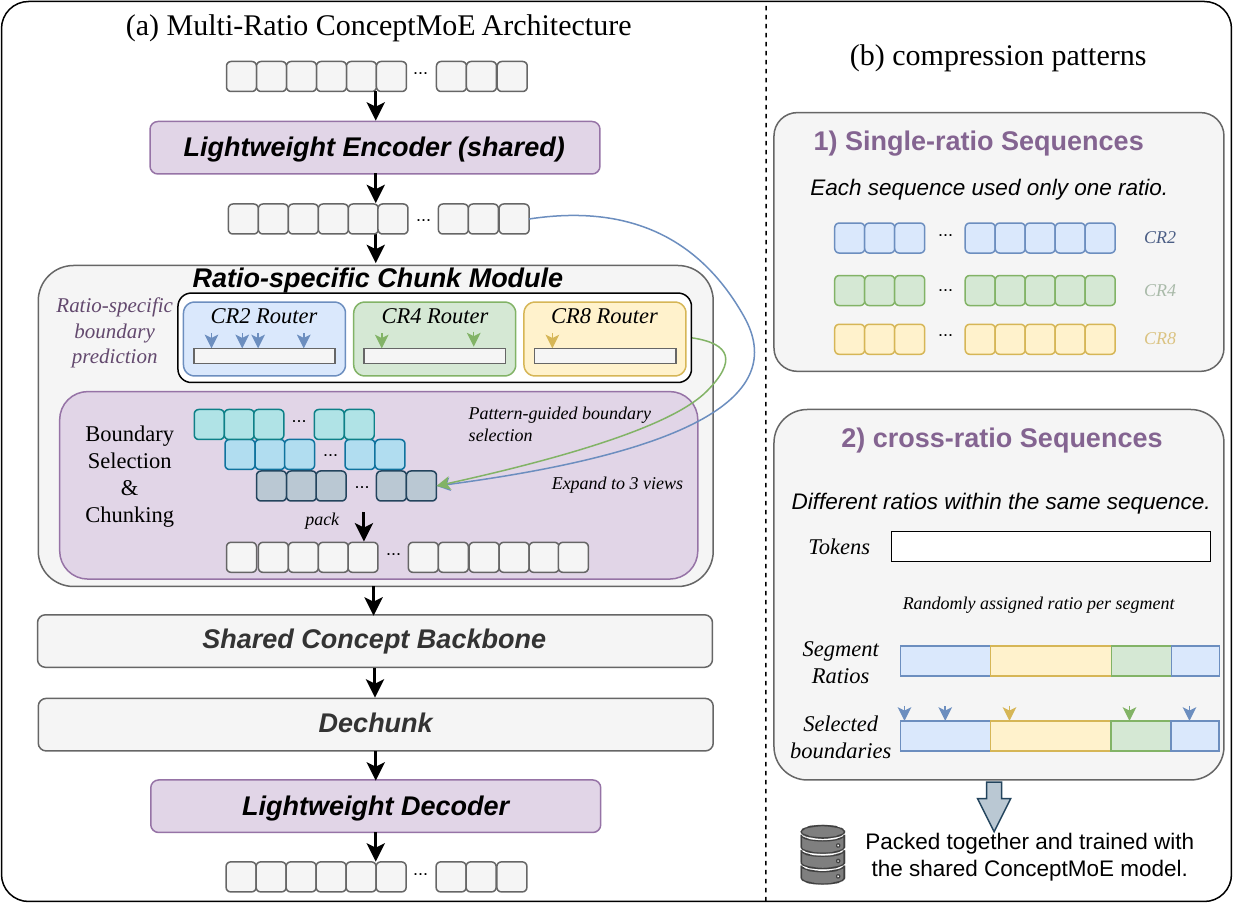}
    \caption{
    \textbf{Overview of Multi-Ratio ConceptMoE.}
    \textbf{(a)} A shared encoder supports ratio-specific boundary prediction
    and multiple compression views, which are packed for shared concept-level
    processing.
    \textbf{(b)} Cross-ratio patterns assign different compression ratios to
contiguous segments, with segment starts forced as chunk boundaries.
    }
    \label{fig:method}
\end{wrapfigure}

To widen the compute range controlled by compression, we first reduce the
compression-independent cost of full-token processing.
We keep the token-level encoder and decoder lightweight and concentrate most
of the model's capacity in the compression-scaled concept backbone.
Specifically, the encoder and decoder retain moderate depth but use a hidden
width of approximately two-thirds of the concept-backbone width.
This reduces $F_{\mathrm{full}}$ in Eq.~\eqref{eq:compute_decomposition},
allowing different compression ratios to produce more clearly separated
compute budgets.

Multi-Ratio ConceptMoE jointly supports three target compression ratios.
For each ratio, the Chunk module contains a lightweight, ratio-specific boundary
router $g_r$:
\begin{equation}
\begin{aligned}
\mathcal{R} &= \{2,4,8\}, \\
\left(P^{(r)},B^{(r)}\right) &= g_r(H),
\qquad r\in\mathcal{R}.
\end{aligned}
\label{eq:ratio_routers}
\end{equation}
where $P^{(r)}\in[0,1]^N$ contains the boundary probabilities and
$B^{(r)}\in\{0,1\}^N$ the corresponding boundary decisions.
All routers operate on the same shared encoder representation $H$, but they are
trained toward different target compression ratios.
Only the boundary routers are ratio-specific; the encoder, concept backbone,
DeChunk module, and decoder are shared across all ratios.

During training, each input is encoded once, and the three routers produce
candidate boundary sequences from the same representation $H$.
We then construct three compression views.
For view $v$, its per-token ratio assignment is represented as
\begin{equation}
A^{(v)}
=
\left(a^{(v)}_1,\ldots,a^{(v)}_N\right)
\in\mathcal{R}^{N},
\label{eq:compression_pattern}
\end{equation}
where $a^{(v)}_i$ specifies the compression ratio assigned to position $i$.
At each position, we select the candidate boundary decision from the router
corresponding to the assigned ratio:
\begin{equation}
\bar{B}^{(v)}_i
=
B^{\left(a^{(v)}_i\right)}_i.
\label{eq:view_boundary_selection}
\end{equation}
These candidate boundaries are combined with the sampled segment boundaries to
obtain the final chunk boundaries, as described in
Section~\ref{sec:cross_ratio_training}.
Figure~\ref{fig:method}(a) summarizes the shared architecture and
ratio-specific boundary selection.

\subsection{Cross-Ratio Compression Training}
\label{sec:cross_ratio_training}

We train Multi-Ratio ConceptMoE with cross-ratio compression patterns that
use different compression ratios within the same sequence.

A single-ratio pattern assigns one compression ratio throughout the sequence,
i.e., $a^{(v)}_i=r$ for all positions.
In contrast, a cross-ratio pattern partitions the sequence into contiguous segments
$\mathcal{I}_1,\ldots,\mathcal{I}_K$ and independently samples a ratio for
each segment:
\begin{equation}
\begin{aligned}
r_k &\sim \mathrm{Uniform}(\mathcal{R}), \\
a^{(v)}_i &= r_k, \qquad i\in\mathcal{I}_k .
\end{aligned}
\label{eq:cross_ratio_pattern}
\end{equation}
The resulting assignment is piecewise constant, as illustrated in
Figure~\ref{fig:method}(b).

To prevent concepts from spanning sampled segments, we force a chunk boundary
at every segment start.
Let $s^{(v)}_i\in\{0,1\}$ indicate whether position $i$ starts a segment.
The final boundary decisions are
\begin{equation}
B^{(v)}_i
=
\bar{B}^{(v)}_i
\lor
s^{(v)}_i,
\qquad i=1,\ldots,N.
\label{eq:forced_boundary}
\end{equation}
All other boundaries within each segment remain content-adaptive and are
predicted by the corresponding ratio-specific router.
The final boundaries are used to construct
$C^{(v)}=\mathrm{Chunk}(H,B^{(v)})$.
The resulting concept sequences are packed for joint processing by the shared
concept backbone while preserving their sequence boundaries.
After concept-level processing, the outputs are separated by view, mapped back
to token resolution through DeChunk, and processed by the shared decoder.

In our final configuration, all three compression views, including their
segmentations and ratio assignments, are independently sampled from the
cross-ratio pattern distribution:
\begin{equation}
A^{(1)},A^{(2)},A^{(3)}
\overset{\mathrm{i.i.d.}}{\sim}
\mathcal{P}_{\mathrm{cross}}.
\label{eq:cross_ratio_views}
\end{equation}
We introduce no explicit cross-ratio distillation, representation alignment,
or consistency objective.
Section~\ref{sec:cross_ratio_ablation} studies alternative mixtures of single-ratio and
cross-ratio training views.

At inference time, an external policy selects a ratio-specific router for each
segment, enabling fixed- or mixed-ratio operation.
Each new segment starts at a forced chunk boundary, so changing its ratio does
not repartition previously constructed concepts.
Since all ratios share the same concept backbone, previously computed history
states remain valid across ratio changes.
\section{Experiments}
\label{sec:experiments}

\subsection{Experimental Setup}
\label{sec:exp_setup}

\paragraph{Models and evaluation.}
We evaluate Multi-Ratio ConceptMoE at three fixed operating points, CR2, CR4, and CR8, against theoretically compute-matched Large, Medium, and Small MoE baselines, as well as independently trained SingleCR specialists.
General capability is evaluated with PTBench on the PT checkpoints and
SFTBench after supervised fine-tuning.
We further evaluate mixed-ratio inference on two multi-turn tasks,
CodeChat~\cite{guo2026codechat} and CoSQL~\cite{yu2019cosql}.
For Pareto analysis, we use Combined Joint accuracy for CodeChat and SQLite
execution accuracy for CoSQL.
Dataset composition, evaluation splits, prompting details, and full evaluation
protocols are provided in Appendix~\ref{app:evaluation}.
For mixed-ratio evaluation, turn-wise CR assignments are generated by
task-specific routing prompts; details are provided in
Appendix~\ref{app:routing_protocol}.

\paragraph{Compute accounting.}
Following the compute-matching setup of ConceptMoE~\cite{DBLP:journals/corr/abs-2601-21420}, the Large, Medium, and
Small MoE baselines are configured to approximately match the activated FLOPs
of CR2, CR4, and CR8, respectively, excluding attention-map computation,
whose cost depends on sequence length and therefore varies with the compression
ratio.
For multi-turn evaluations, we report workload-realized FLOPs under a shared
terminal-turn workload, including attention-map and sliding-window-attention
computation.
The same prompt and output-token workload is used across models, while
ConceptMoE FLOPs are computed using the realized prefill and decode compression
ratios.
Full accounting details are provided in Appendix~\ref{app:flops}.

\subsection{Strong Fixed-Ratio Operating Points}
\label{sec:fixed_ratio_results}
%
%

\begin{figure}[t]
\centering

\begin{minipage}[t]{0.485\linewidth}
\centering
\captionof{table}{
\textbf{Against compute-matched MoE baselines.}
Each cell reports MoE / Multi-Ratio accuracy (\%).
Multi-Ratio ConceptMoE wins all 6 comparisons.
}
\label{tab:fixed_vs_moe}
\small
\setlength{\tabcolsep}{6pt}
\begin{tabular}{lcc}
\toprule
Budget
& PTBench
& SFTBench \\
\midrule
High
& 31.7 / \textbf{32.1}
& 19.0 / \textbf{21.5} \\
Medium
& 28.2 / \textbf{29.3}
& 15.5 / \textbf{19.1} \\
Low
& 25.4 / \textbf{27.6}
& 13.1 / \textbf{17.2} \\
\bottomrule
\end{tabular}
\end{minipage}
\hfill
\begin{minipage}[t]{0.485\linewidth}
\centering
\captionof{table}{
\textbf{Against independently trained specialists.}
Each cell reports SingleCR / Multi-Ratio accuracy (\%).
}
\label{tab:fixed_vs_specialist}
\small
\setlength{\tabcolsep}{6pt}
\begin{tabular}{lcc}
\toprule
Ratio
& PTBench
& SFTBench \\
\midrule
CR2
& \textbf{33.2} / 32.1
& 21.1 / \textbf{21.5} \\
CR4
& \textbf{30.0} / 29.3
& 18.8 / \textbf{19.1} \\
CR8
& 25.8 / \textbf{27.6}
& 16.4 / \textbf{17.2} \\
\midrule
Avg.
& 29.7 / 29.7
& 18.8 / \textbf{19.3} \\
\bottomrule
\end{tabular}
\end{minipage}

\end{figure}

\begin{table}[t]
\centering
\caption{
\textbf{Accuracy--compute trade-offs on CodeChat and CoSQL.}
CodeChat reports Combined Joint accuracy, while CoSQL reports SQLite
execution accuracy. FLOPs include attention-map computation and use the
terminal-turn workload protocol.
}
\label{tab:pareto}
\begin{tabular}{lrr|rr}
\toprule
& \multicolumn{2}{c|}{CodeChat}
& \multicolumn{2}{c}{CoSQL} \\
Model & Acc. (\%) & FLOPs (PF) & Acc. (\%) & FLOPs (PF) \\
\midrule
MoE Large  & 37.91 & 1.8662 & 49.16 & 0.3785 \\
MoE Medium & 27.09 & 1.1198 & 25.52 & 0.2283 \\
MoE Small  & 18.68 & 0.7334 & 14.00 & 0.1502 \\
\midrule
Multi-Ratio CR2   & 43.28 & 1.3072 & \textbf{54.72} & 0.3788 \\
Multi-Ratio CR4   & 35.37 & 0.9529 & 48.26 & 0.2418 \\
Multi-Ratio CR8   & 29.29 & 0.7795 & 41.11 & 0.1863 \\
\textbf{Multi-Ratio Mixed}
& \textbf{43.78} & 1.0999 & 51.84 & 0.3319 \\
\midrule
SingleCR2 & 33.89 & 2.2290 & 54.02 & 0.4027 \\
SingleCR4 & 24.60 & 1.2784 & 42.50 & 0.2873 \\
SingleCR8 & 18.02 & 0.7326 & 33.37 & 0.2427 \\
\bottomrule
\end{tabular}
\end{table}

\begin{table}[t]
\centering
\small
\caption{
\textbf{Switching overhead under the same turn-wise compute-budget schedule.}
For the routed independent-model baseline, total FLOPs include the additional
history re-prefill required after model switches.
The comparison quantifies the algorithmic compute overhead of switching among
independent models.
}
\label{tab:model_switching}
\begin{tabular}{lrrrrr}
\toprule
Dataset
& Switches
& \shortstack{Re-prefilled\\Tokens}
& \shortstack{Routed\\FLOPs (PF)}
& \shortstack{Multi-Ratio\\FLOPs (PF)}
& Reduction \\
\midrule
CodeChat
& 1,925
& 2,882,580
& 5.0895
& 1.0999
& 78.4\% \\
CoSQL
& 512
& 320,108
& 0.6250
& 0.3319
& 46.9\% \\
\bottomrule
\end{tabular}
\end{table}

\begin{table}[t]
\centering
\caption{
\textbf{Effect of cross-ratio training exposure.}
We compare homogeneous-only training (3 single-ratio views),
mixed training (1 single-ratio + 2 cross-ratio views), and fully
cross-ratio training (3 cross-ratio views).
PTBench and SFTBench report average fixed-ratio accuracy across CR2/CR4/CR8,
while CodeChat and CoSQL report mixed-ratio inference accuracy.
}
\label{tab:cross_ratio_ablation}
\setlength{\tabcolsep}{6.5pt}
\begin{tabular}{lcc|cc}
\toprule
Training composition
& \multicolumn{2}{c|}{Fixed-Ratio Evaluation}
& \multicolumn{2}{c}{Mixed-Ratio Inference} \\
& PTBench
& SFTBench
& CodeChat
& CoSQL \\
\midrule

3 Single + 0 Cross
& \textbf{29.97}
& 18.83
& 37.65
& 43.00 \\

1 Single + 2 Cross
& \textbf{29.97}
& 18.73
& 37.57
& 45.88 \\

0 Single + 3 Cross
& 29.67
& \textbf{19.27}
& \textbf{43.78}
& \textbf{51.84} \\

\bottomrule
\end{tabular}
\end{table}

We first examine whether a single Multi-Ratio checkpoint can provide strong
standalone operating points across compute budgets.
As shown in Table~\ref{tab:fixed_vs_moe}, CR2, CR4, and CR8 all outperform
their corresponding Large, Medium, and Small MoE baselines on both PTBench and SFTBench, yielding \textbf{6 wins out of 6 comparisons}.
The gains span all three evaluated compute budgets, showing that the shared
checkpoint retains strong fixed-ratio performance across the evaluated compute
range.

We further compare Multi-Ratio ConceptMoE with independently trained
SingleCR specialists.
Although the specialists retain small advantages on several PT evaluations,
the shared model remains competitive overall.
Averaged across CR2/CR4/CR8, Multi-Ratio matches the independently trained
specialists on PTBench at 29.7 and improves SFTBench from 18.8 to 19.3.
The advantage is particularly clear at CR8, where Multi-Ratio outperforms
the corresponding specialist on both evaluation suites.
These results indicate that supporting multiple compression modes in one
shared model does not introduce a substantial overall specialization penalty.

\subsection{Mixed-Ratio Inference Improves the Accuracy--Compute Trade-off}
\label{sec:mixed_pareto}

We next evaluate whether composing different compression modes across turns
can improve the accuracy--compute trade-off.
Figure~\ref{fig:teaser}(b,c) and Table~\ref{tab:pareto} show that Mixed lies
on the accuracy--compute Pareto frontier of the compared configurations on
both CodeChat and CoSQL.
The comparison includes MoE baselines, fixed CR2/CR4/CR8 modes, and
independently trained SingleCR specialists.

On CodeChat, Mixed achieves \textbf{43.78\%} accuracy at
\textbf{1.0999 PFLOPs}, outperforming the Large MoE by 5.87 points while
using 41.1\% fewer FLOPs, and strictly dominating fixed CR2.
On CoSQL, Mixed reaches \textbf{51.84\%} execution accuracy at
\textbf{0.3319 PFLOPs}, improving over the Large MoE by 2.68 points with
12.3\% fewer FLOPs.
Compared with fixed CR2, it sacrifices only 2.88 points of accuracy while
reducing compute by 12.4\%.
Thus, on both evaluated multi-turn tasks, turn-wise composition yields
accuracy--compute operating points that are not provided by the evaluated
fixed-budget alternatives.

\subsection{Switching Overhead under the Same Budget Schedule}
\label{sec:model_switching}

A natural alternative to mixed-ratio inference is to route each turn among
separately deployed Large, Medium, and Small models.
To quantify the compute overhead associated with model switching, we apply exactly
the same frozen turn-wise budget schedule used by Multi-Ratio ConceptMoE
(Appendix~\ref{app:routing_protocol}), mapping CR2/CR4/CR8 to
Large/Medium/Small, respectively.

\begin{figure}[t]
    \centering
    \includegraphics[width=0.86\linewidth]{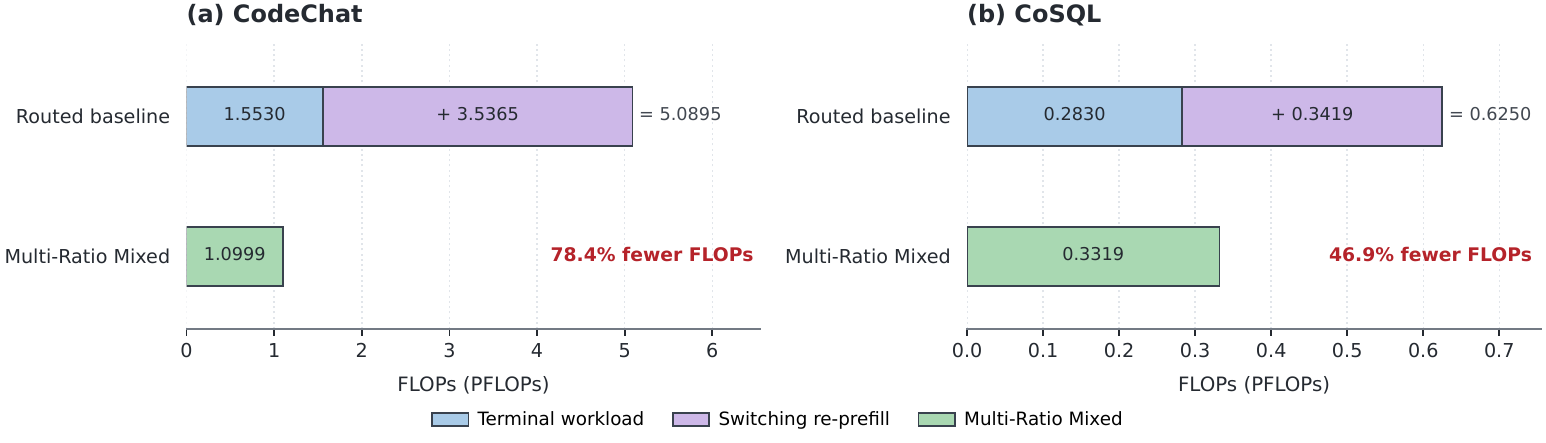}
    \caption{
    \textbf{FLOPs breakdown for turn-wise compute allocation.}
    Both approaches follow the same turn-wise compute-budget schedule.
    Routing among independent models incurs additional history re-prefill
    whenever the selected model changes, whereas Multi-Ratio ConceptMoE
    changes compression modes within a shared model.
    }
    \label{fig:switching_flops}
\end{figure}

Both approaches use the same terminal-turn workload convention.
Multi-Ratio ConceptMoE changes compression modes within one shared model,
whereas independently deployed models maintain model-specific history states.
We therefore account for the history re-prefill required to reconstruct the
state of the newly selected independent model after each switch.

Figure~\ref{fig:switching_flops} visualizes the resulting overhead, and
Table~\ref{tab:model_switching} reports the corresponding FLOPs.
On CodeChat, the frozen schedule contains 1,925 model switches and requires
approximately 2.88M history tokens to be re-prefilled.
This adds 3.5365 PFLOPs of history re-prefill to the routed baseline,
increasing its total computation from 1.5530 to 5.0895 PFLOPs.
Multi-Ratio Mixed uses 1.0999 PFLOPs in total, 78.4\% lower than the
resulting routed total.
On CoSQL, 512 switches require approximately 320K re-prefilled history tokens,
raising the routed baseline from 0.2830 to 0.6250 PFLOPs, while Multi-Ratio
Mixed uses 0.3319 PFLOPs, corresponding to a 46.9\% reduction.

This comparison is intended to quantify the algorithmic compute overhead
introduced by switching among independent models.

\subsection{Effect of Cross-Ratio Training Exposure}
\label{sec:cross_ratio_ablation}

\begin{figure}[t]
    \centering
    \includegraphics[width=0.90\linewidth]{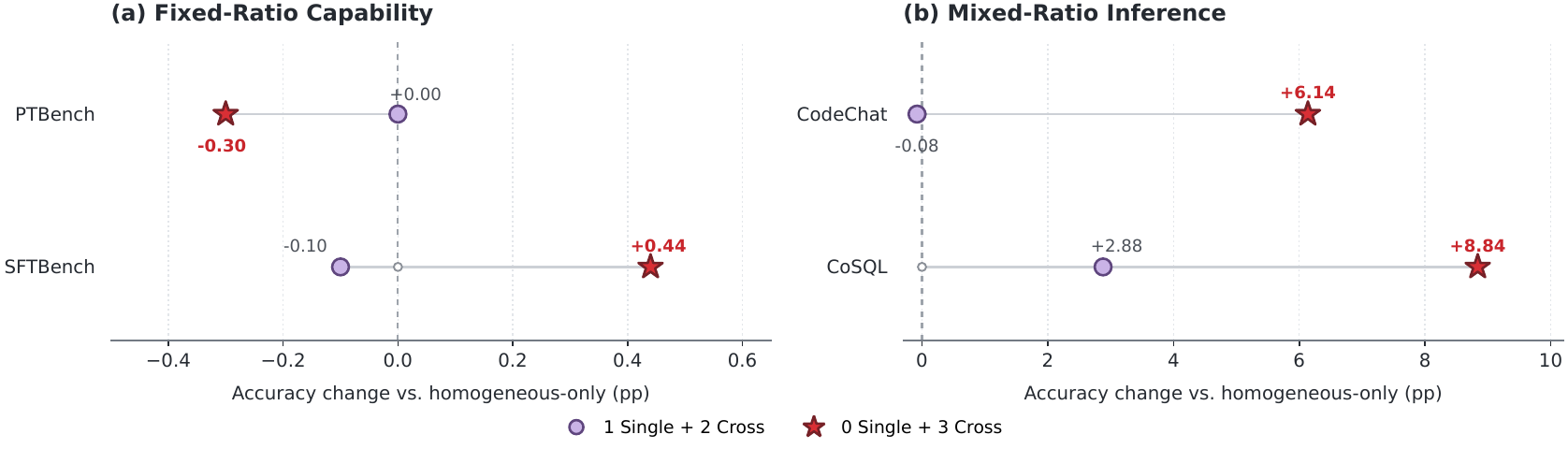}
    \caption{
    \textbf{Effect of cross-ratio training exposure.}
    Accuracy changes are measured relative to homogeneous-only training
    (3 Single + 0 Cross).
    Cross-ratio exposure largely preserves fixed-ratio capability, while
    fully cross-ratio training substantially improves mixed-ratio inference
    on CodeChat and CoSQL.
    }
    \label{fig:cross_ratio_exposure}
\end{figure}

Finally, we study whether cross-ratio training exposure is important for
composing different compression modes.
We compare three training compositions: three single-ratio views,
one single-ratio plus two cross-ratio views, and the fully cross-ratio
configuration with three cross-ratio views.
Figure~\ref{fig:cross_ratio_exposure} visualizes the accuracy changes relative
to homogeneous-only training, while Table~\ref{tab:cross_ratio_ablation}
reports the corresponding absolute scores.

For fixed-ratio general capability, the effect of cross-ratio exposure is
relatively small.
Compared with homogeneous-only training, the fully cross-ratio configuration
changes PTBench by $-0.30$ points, while
improving SFTBench by $0.44$ points.
The intermediate configuration with one single-ratio and two cross-ratio
views likewise remains close to the homogeneous baseline.
Thus, exposing all training views to cross-ratio patterns does not cause a
substantial overall loss in the standalone utility of the fixed operating
points.

The effect is much larger for mixed-ratio inference.
Relative to homogeneous-only training, fully cross-ratio training improves
CodeChat by \textbf{6.14} points and CoSQL by \textbf{8.84} points, while
the corresponding inference FLOPs remain comparable.
Overall, fully cross-ratio training largely preserves fixed-ratio capability
and slightly improves it after SFT, while substantially strengthening the
ability to compose different compression modes.
We therefore use the fully cross-ratio configuration as our final model.

\section{Conclusion}
\label{sec:conclusion}

We study flexible compute allocation across turns without switching among
separately deployed models.
To this end, we introduce Multi-Ratio ConceptMoE, which exposes multiple
compression-based operating points within a shared model.
By keeping full-token processing lightweight, using ratio-specific boundary routing, and training with cross-ratio compression patterns, a single
checkpoint can operate at fixed CR2/CR4/CR8 budgets or compose these modes
across turns.
Across general evaluations, the fixed modes consistently outperform their
compute-matched MoE baselines, while remaining competitive with independently
trained fixed-ratio specialists.
On CodeChat and CoSQL, mixed-ratio inference improves the accuracy--compute
trade-off and avoids the cross-model state
reconstruction overhead accounted for in our routed baseline.
These results demonstrate that a single shared model can support flexible
compute allocation across turns without cross-model switching.

\clearpage

\bibliographystyle{plainnat}
\bibliography{main}

\clearpage

\beginappendix

\section{Detailed Model and Training Setup}
\label{app:model_training}

\subsection{Multi-Ratio Configuration}
\label{app:multi_ratio_config}

Multi-Ratio ConceptMoE supports three target compression ratios,
\[
\mathcal{R}=\{2,4,8\}.
\]
Each ratio uses an independent lightweight boundary router, while the encoder,
concept backbone, DeChunk module, and decoder are shared across all ratios.
The encoder and decoder use approximately two-thirds of the hidden width of
the concept backbone to reduce full-token computation.

During training, each input is encoded once and expanded into three
compression views from the same encoder representation.
The views are chunked independently according to their sampled compression
patterns, packed for shared concept-level processing, and separated again
before DeChunk and decoding.
Unless otherwise specified, our final model uses three independently sampled
cross-ratio views.

\subsection{Baselines}
\label{app:baselines}

We use Large, Medium, and Small MoE baselines for the high-, medium-, and
low-compute operating points, respectively.
Their non-attention activated FLOPs are approximately matched to CR2, CR4,
and CR8:
\[
\mathrm{Large}\leftrightarrow\mathrm{CR2},\qquad
\mathrm{Medium}\leftrightarrow\mathrm{CR4},\qquad
\mathrm{Small}\leftrightarrow\mathrm{CR8}.
\]

We additionally train three independently optimized fixed-ratio ConceptMoE
specialists, denoted as SingleCR2, SingleCR4, and SingleCR8.
For the cross-ratio training ablation, we compare
\[
3~\mathrm{single}+0~\mathrm{cross},\qquad
1~\mathrm{single}+2~\mathrm{cross},\qquad
0~\mathrm{single}+3~\mathrm{cross}.
\]

\subsection{Cross-Ratio Training Details}
\label{app:cross_ratio_details}

For each cross-ratio view, a training sequence is partitioned into contiguous
token segments.
Interior segment lengths are sampled uniformly from 64 to 960 tokens, with an
expected length of 512 tokens; a residual suffix shorter than 64 tokens is
absorbed into the preceding segment.
The segmentation is performed at the token level and is not constrained to
align with semantic or turn boundaries.
Each segment independently samples a compression ratio,
\[
r_k \sim \mathrm{Uniform}\{2,4,8\}.
\]
The assigned ratio remains constant within a segment.
At every segment start, we force a chunk boundary to keep concept formation
aligned with the sampled segmentation, while all other boundaries are predicted
adaptively by the corresponding ratio-specific router.
This procedure defines the cross-ratio pattern distribution
$\mathcal{P}_{\mathrm{cross}}$ used in Section~\ref{sec:cross_ratio_training}.

The three compression views are sampled independently, so the same encoder
representation can be associated with different segmentations and ratio
assignments within the same training step.
After pattern-guided boundary selection and chunking, the resulting concept
sequences are packed and processed jointly by the shared concept backbone.

Boundary prediction, hard boundary construction, and the compression-control
objective follow ConceptMoE~\cite{DBLP:journals/corr/abs-2601-21420}
without modification.
For cross-ratio training, the compression-control statistics of each
ratio-specific router are computed only over tokens assigned to that ratio.
Training otherwise uses the standard token-level language-modeling objective
together with the inherited compression-control auxiliary objective.
We introduce no additional cross-ratio distillation, representation alignment,
or consistency objective.

The same Multi-Ratio architecture is used for both PT and SFT.
Unless otherwise specified, all final results use three independently sampled
cross-ratio compression views per input.










\begin{table}[t]
\centering
\small
\caption{
\textbf{Evaluation sets included in PTBench across different domains.}
}
\label{tab:ptbench_details}

\renewcommand{\arraystretch}{1.08}
\setlength{\tabcolsep}{4pt}

\begin{tabularx}{\linewidth}{
>{\raggedright\arraybackslash}X
>{\raggedright\arraybackslash}X
>{\raggedright\arraybackslash}X
>{\raggedright\arraybackslash}X
>{\raggedright\arraybackslash}X
>{\raggedright\arraybackslash}X
}
\toprule

\centering\arraybackslash\textbf{\shortstack{Comprehensive\\Evaluation}} &
\centering\arraybackslash\textbf{Reasoning} &
\centering\arraybackslash\textbf{Math} &
\centering\arraybackslash\textbf{Code} &
\centering\arraybackslash\textbf{Knowledge} &
\centering\arraybackslash\textbf{\shortstack{Specialized\\Knowledge}}
\\
\midrule

\begin{minipage}[t]{\linewidth}\raggedright
MMLU~\cite{hendrycks2020measuring}\par
C-Eval~\cite{huang2023c}\par
MMLU-Pro~\cite{wang2024mmlu}\par
AGIEval~\cite{zhong2024agieval}\par
Humanity's Last Exam~\cite{phan2025humanity}\par
SuperGPQA~\cite{du2026supergpqa}\par
LiveBench~\cite{white2025livebench}
\end{minipage}
&
\begin{minipage}[t]{\linewidth}\raggedright
BBH~\cite{suzgun2023challenging}\par
DROP~\cite{dua2019drop}\par
ARC-AGI~\cite{chollet2019measure}\par
ProcBench~\cite{fujisawa2024procbench}\par
ZebraLogic~\cite{lin2025zebralogic}\par
KOR-Bench~\cite{ma2025kor}
\end{minipage}
&
\begin{minipage}[t]{\linewidth}\raggedright
MATH~\cite{hendrycks2021measuring}\par
AIME 2024\par
AIME 2025\par
HARP~\cite{yue2024harp}\par
Omni-MATH~\cite{gao2025omni}\par
OlympiadBench~\cite{he2024olympiadbench}
\end{minipage}
&
\begin{minipage}[t]{\linewidth}\raggedright
HumanEval~\cite{chen2021evaluating}\par
MBPP+~\cite{austin2021program,liu2023your}\par
LiveCode Bench~\cite{jain2025livecodebench}\par
McEval~\cite{chai2025mceval}
\end{minipage}
&
\begin{minipage}[t]{\linewidth}\raggedright
TriviaQA~\cite{joshi2017triviaqa}\par
SimpleQA~\cite{wei2024measuring}\par
Chinese SimpleQA~\cite{he2025chinese}
\end{minipage}
&
\begin{minipage}[t]{\linewidth}\raggedright
GPQA~\cite{rein2023gpqa}
\end{minipage}
\\

\bottomrule
\end{tabularx}
\end{table}











\begin{table}[t]
\centering
\footnotesize
\caption{
\textbf{Evaluation sets included in SFTBench across different domains.}
}
\label{tab:sftbench_details}

\renewcommand{\arraystretch}{1.06}
\setlength{\tabcolsep}{2.2pt}

\begin{tabularx}{\linewidth}{
>{\raggedright\arraybackslash}X
>{\raggedright\arraybackslash}X
>{\raggedright\arraybackslash}X
>{\raggedright\arraybackslash}X
>{\raggedright\arraybackslash}X
>{\raggedright\arraybackslash}X
>{\raggedright\arraybackslash}X
}
\toprule

\centering\arraybackslash
\textbf{\shortstack{Comprehensive\\Evaluation}}
&
\centering\arraybackslash
\textbf{Reasoning}
&
\centering\arraybackslash
\textbf{Math}
&
\centering\arraybackslash
\textbf{Code}
&
\centering\arraybackslash
\textbf{\shortstack{Instruction\\Following}}
&
\centering\arraybackslash
\textbf{\shortstack{Knowledge /\\Hallucination}}
&
\centering\arraybackslash
\textbf{Multilingual}
\\
\midrule

\begin{minipage}[t]{\linewidth}\raggedright
MMLU-Pro~\cite{wang2024mmlu}\par
MMLU~\cite{hendrycks2020measuring}\par
Humanity's Last Exam~\cite{phan2025humanity}\par
GPQA Diamond~\cite{rein2023gpqa}\par
SuperGPQA~\cite{du2026supergpqa}\par
LiveBench~\cite{white2025livebench}\par
MixEvalHard~\cite{ni2024mixeval}\par
ArenaHard~\cite{li2024crowdsourced}
\end{minipage}
&
\begin{minipage}[t]{\linewidth}\raggedright
ARC-AGI~\cite{chollet2019measure}\par
ProcBench~\cite{fujisawa2024procbench}\par
KOR-Bench~\cite{ma2025kor}\par
ZebraLogic Bench~\cite{lin2025zebralogic}
\end{minipage}
&
\begin{minipage}[t]{\linewidth}\raggedright
AIME 2025\par
AIME 2024\par
HARP~\cite{yue2024harp}\par
Omni-MATH~\cite{gao2025omni}\par
OlympiadBench~\cite{he2024olympiadbench}
\end{minipage}
&
\begin{minipage}[t]{\linewidth}\raggedright
Aider\par
LiveCode Bench~\cite{jain2025livecodebench}
\end{minipage}
&
\begin{minipage}[t]{\linewidth}\raggedright
MultiChallenge~\cite{deshpande2025multichallenge}\par
IFEval~\cite{zhou2023instruction}\par
Collie Hard
\end{minipage}
&
\begin{minipage}[t]{\linewidth}\raggedright
Chinese SimpleQA~\cite{he2025chinese}\par
SimpleQA~\cite{wei2024measuring}
\end{minipage}
&
\begin{minipage}[t]{\linewidth}\raggedright
MMMLU~\cite{singh2025global}
\end{minipage}
\\

\bottomrule
\end{tabularx}
\end{table}

\section{Evaluation Protocols}
\label{app:evaluation}

\subsection{General Benchmarks}
\label{app:general_benchmarks}

We use two aggregate evaluation suites for different training stages:
PTBench for PT checkpoints and SFTBench for supervised fine-tuned checkpoints.
Their benchmark compositions are summarized in
Tables~\ref{tab:ptbench_details} and~\ref{tab:sftbench_details}, respectively.

PTBench and SFTBench aggregate their category-level scores using fixed
category weights.
For PTBench, the weights for Comprehensive, Reasoning, Math, Code,
Specialized Knowledge, and Knowledge are
$0.30/0.15/0.15/0.15/0.10/0.15$, respectively.
For SFTBench, the weights for Comprehensive, Reasoning, Math, Code,
Instruction Following, Knowledge, and Multilingual are
$0.35/0.10/0.15/0.15/0.10/0.10/0.05$, respectively.

\subsection{CodeChat-Eval}
\label{app:codechat_protocol}

We use the MBPP+-based subset of CodeChat-Eval~\cite{austin2021program, liu2023your} consisting of
378 multi-turn code-refinement trajectories. Each trajectory contains one initial programming turn followed by nine
refinement turns, yielding 3,780 turns in total, including 3,402 refinement
turns.
Among them, 3,402 are refinement turns.

We report three complementary metrics.

\paragraph{Functional.}
For every turn, we use EvalPlus+ functional tests to determine whether the
generated program is functionally correct.

\paragraph{Refinement Joint.}
For turns 1--9, a prediction is considered correct only when it satisfies
both functional correctness and the corresponding refinement instruction:
\[
\mathrm{Joint}
=
\mathrm{Functional}
\land
\mathrm{Instruction\ Adherence}.
\]
Functional correctness is determined by EvalPlus+ tests, while instruction
adherence is judged by Terra based on GPT-5.6.
Refinement Joint is computed over all 3,402 refinement turns and measures
whether a model can perform the requested edit while preserving program
functionality.

\paragraph{Combined Joint.}
To obtain a single metric over complete trajectories, turn 0 is evaluated
using functional correctness, while turns 1--9 use the Joint criterion:
\[
\mathrm{CombinedJoint}
=
\frac{
N_{\mathrm{functional},\,t=0}
+
N_{\mathrm{joint},\,t=1:9}
}{
3780
}.
\]
We use Combined Joint for the accuracy--compute Pareto analysis in the main
paper, while Refinement Joint is additionally reported as a more direct
measure of code-editing capability.

\subsection{CoSQL}
\label{app:cosql_protocol}

We evaluate conversational text-to-SQL on the development split of
CoSQL~\cite{yu2019cosql}.
Our primary metric is turn-level SQLite execution accuracy.

For each turn, we execute both the predicted SQL and the gold SQL against
the corresponding database and compare their query results.
If the gold query explicitly contains an \texttt{ORDER BY} clause, result
order is preserved during comparison; otherwise, row order is ignored.
A prediction is counted as incorrect if the SQL cannot be parsed or
executed, if generation terminates before a complete query is produced, or
if its execution result differs from that of the gold query.

We additionally compute Exact Match as an auxiliary diagnostic metric.
However, syntactically different SQL queries can be semantically equivalent
and yield identical execution results.
We therefore use execution accuracy, rather than Exact Match, as the primary
CoSQL metric throughout the main paper.

\subsection{Prompting and Generation Protocol}
\label{app:prompting}

Different checkpoints may require different prompting or prefill formats for
stable and well-formed generation, particularly when they are sensitive to
special reasoning prefixes such as \texttt{<think>}.
We therefore use checkpoint-specific compatible configurations to avoid
formatting or truncation artifacts, while keeping the task instructions and
evaluation criteria unchanged within each benchmark.

\subsection{Mixed-Ratio Routing Protocol}
\label{app:routing_protocol}

For mixed-ratio evaluation, we construct task-specific turn-wise compute
schedules using \texttt{gpt-5.6-terra} as the routing judge.
Each decision is generated from a physically prefix-isolated record containing
only information available through the current turn. The resulting schedules
are frozen before evaluating the reported Multi-Ratio checkpoint.
The router does not observe model outputs, correctness judgments, evaluation
results, or future turns.
The same frozen routing schedule is used when evaluating Multi-Ratio ConceptMoE
and the corresponding routed independent-model baseline.

\paragraph{CodeChat-Eval.}
For each code-refinement turn, the router receives a separate record containing
only the original programming problem, the refinement instructions up to and
including the current turn, and the current turn index.
It does not observe model-generated code, functional-test outcomes,
instruction-adherence judgments, fixed-ratio results, future refinement
instructions, or model identity.
The router is instructed to select the lowest-cost compression mode that
maintains an acceptable risk of preserving the behavioral constraints visible
so far.

The exact routing prompt is:

{\footnotesize
\begin{verbatim}
You are a serving-time routing judge for one turn of a cumulative code-editing
conversation.

Produce three separate judgments:

1. a raw inference tier: CR2, CR4, or CR8;
2. a risk class;
3. confidence in the interpretation of the current request.

The raw tier is an initial routing judgment. A separate deterministic quality
guard will later promote turn 0, all `behavioral` turns, and all low-confidence
turns to CR2. Do not perform that later promotion yourself merely because a
turn is behavioral or low-confidence.

The input is physically prefix-isolated. Use only the original problem,
`visible_instructions` ending at the current turn, `current_instruction`, and
`turn_index`. No future instruction, generated code, test, EvalPlus result,
instruction-following judgment, fixed-CR result, catalog category, or prior
router label is available.

Risk class:

- `mechanical_local`: a clearly local edit, such as a rename, comment,
  formatting/presentation change, or narrowly specified syntactic rewrite.
- `stateful_refactor`: a bounded implementation/refactor that must preserve the
  visible accumulated requirements but does not clearly request changed
  behavior.
- `behavioral`: an edit that changes or may change behavior, invariants,
  algorithm, data structures, error/boundary handling, concurrency, or resource
  semantics.

For turn 0, classify an ordinary bounded initial implementation as
`stateful_refactor`; use `behavioral` only when the original problem itself has
substantial algorithmic, invariant, boundary-condition, or ambiguity risk. Do
not call every initial implementation behavioral merely because it creates
program behavior.

Confidence describes how unambiguously the visible prefix determines the
intended edit. An underspecified instruction can be `low` confidence even when
its risk class is `stateful_refactor`. Do not change the risk class merely to
encode confidence.

Raw CR policy:

- CR8 only for `mechanical_local` with high confidence.
- CR4 is the default for an ordinary bounded implementation or edit, including
  many stateful refactors, underspecified edits, and behavioral edits whose
  raw need for exact context is not clearly critical.
- CR2 only when exact high-fidelity context is already clearly critical before
  the later quality guard: extensive accumulated constraints, fragile
  long-distance state, several interacting edits, or a particularly complex
  behavior/invariant change.
- Turn 0 normally uses CR4; use CR2 only for an unusually complex or fragile
  initial implementation. Never use CR8 for turn 0.

Do not force CR2 solely because `risk_class` is `behavioral` or confidence is
`low`; the later quality guard handles those conservative promotions.

Return exactly one JSON object per input record:

{
  "route_key": "Mbpp/2::turn-3",
  "task_id": "Mbpp/2",
  "turn_index": 3,
  "input_sha256": "...",
  "route_cr": 4,
  "risk_class": "stateful_refactor",
  "confidence": "low",
  "reason": "one short concrete reason",
  "router": "terra_codechat_strict_causal_legacy_aligned_v2"
}

Allowed risk classes are `behavioral`, `stateful_refactor`, and
`mechanical_local`. Confidence must be `low`, `medium`, or `high`. Preserve
identity and hash fields exactly. Return JSONL only, in input order.
\end{verbatim}
}

For the reported CodeChat-Eval results, we additionally apply a deterministic
quality guard to the raw router outputs.
The initial implementation is always assigned CR2; behavioral-risk edits and
low-confidence decisions are also promoted to CR2:
\begin{equation}
r_t =
\begin{cases}
2, & t=0, \\
2, & \text{risk}_t=\texttt{behavioral}, \\
2, & \text{confidence}_t=\texttt{low}, \\
r^{\mathrm{Terra}}_t, & \text{otherwise}.
\end{cases}
\label{eq:codechat_route_guard}
\end{equation}
Thus, CR8 is reserved for high-confidence mechanical edits, while CR4 is used
primarily for bounded local refactors.

\paragraph{CoSQL.}
For each user turn, the router receives a separate record containing only the
database schema, user utterances up to and including the current turn, and the
current turn index.
It does not observe gold SQL, model-generated SQL, execution results,
exact-match scores, fixed-ratio correctness, or future utterances.
The routing objective is to select the lowest-cost compression mode with
acceptable risk of corrupting the accumulated SQL state.

The exact user-turn routing prompt is:

{\footnotesize
\begin{verbatim}
You are a routing judge for one turn of a multi-turn text-to-SQL conversation.

Choose exactly one inference tier for the current user turn: CR2, CR4, or CR8.
CR2 is strongest and most expensive; CR8 is cheapest and most compressed.

The input is physically prefix-isolated. You may use only:

1. the database schema;
2. `visible_utterances`, which ends at the current turn;
3. `current_utterance` and `turn_index`.

No future utterance, gold SQL, model-generated SQL, execution result, test
outcome, fixed-CR correctness, or hidden label is available. Do not infer or
invent unseen future requests.

Choose the lowest-cost tier with acceptable risk of corrupting the accumulated
SQL state:

- CR8: only for a clearly local, mechanical, high-confidence operation, such as
  changing an explicit constant, sort direction, limit, or projected field,
  where the referent and prior constraints are unambiguous and no join,
  grouping, query-plan, nesting, or set-logic restructuring is needed.
- CR4: the default for a bounded follow-up that requires the visible dialogue
  state but makes one clear change without substantially restructuring joins,
  grouping, nesting, or set logic.
- CR2: use for the initial query; ambiguous or long-distance references;
  multiple accumulated constraints; corrections or replacement of earlier
  constraints; joins with non-obvious paths; nesting; set operations; negation;
  grouping or HAVING; derived aggregates; or any low-confidence decision where
  an error can corrupt later turns.

The initial user turn must use CR2. Any low-confidence decision must use CR2.
CR8 is allowed only for `mechanical_local` risk with high confidence. If
uncertain between CR4 and CR8, choose CR4. If uncertain whether established SQL
state will be preserved, choose CR2.

For each input JSON object, return exactly one JSON object with:

{
  "route_key": "cosql_dev_0000::turn-0",
  "interaction_id": "cosql_dev_0000",
  "dev_index": 0,
  "turn_index": 0,
  "route_cr": 2,
  "risk_class": "state_establishment",
  "confidence": "high",
  "reason": "one short concrete reason"
}

Allowed risk classes are `state_establishment`, `semantic_state_change`,
`bounded_state_update`, and `mechanical_local`. Preserve the input identity
fields exactly. Return JSONL only, in input order.
\end{verbatim}
}

The unsplit system message, which contains the task instruction and complete
database schema, is routed separately.
We assign CR2 when the schema contains at least four tables or when a
classifier trained only on the CoSQL training split predicts the initial SQL
query to have medium-or-higher hardness; otherwise, we assign CR4:
\begin{equation}
r_{\mathrm{sys}} =
\begin{cases}
2, & \text{complex schema or medium-or-harder predicted query}, \\
4, & \text{otherwise}.
\end{cases}
\label{eq:cosql_system_route}
\end{equation}
The classifier uses only schema statistics and features derived from the
visible user-utterance prefix. For the system decision, the current utterance is
the initial user utterance, so no future utterance is available. CoSQL
development-set gold SQL and evaluation outcomes are not used.

Within each benchmark, fixed-ratio and mixed-ratio runs use the same generation
prompt and decoding parameters; only the compression-ratio assignment changes.

For completeness, the frozen CodeChat schedule assigns 2,125 (56.2\%),
1,062 (28.1\%), and 593 (15.7\%) turns to CR2, CR4, and CR8, respectively,
and contains 1,925 transitions between compression modes.

For CoSQL, the 293 system messages are assigned to CR2/CR4 in 193/100 cases,
while the 1,007 user turns are assigned to CR2/CR4/CR8 in 539/334/134 cases,
respectively. The resulting CoSQL schedules contain 512 compression-mode
transitions, including transitions from the system message to the first user
turn.

The same frozen routing schedules are used across the cross-ratio training
ablations, so the differences in mixed-ratio accuracy arise under an unchanged
external compute-allocation policy.

\section{FLOPs Accounting}
\label{app:flops}

\subsection{Terminal-Turn Workload}
\label{app:terminal_flops}

For CodeChat and CoSQL, we use a standardized \emph{terminal-turn workload}
for multi-turn FLOPs evaluation, decomposed into prefill and decode costs:
\[
F_{\mathrm{terminal}}
=
F_{\mathrm{prefill}}
+
F_{\mathrm{decode}}.
\]

Within each benchmark, all models use matched prompt and output token
workloads, enabling direct comparison of computational cost across model
architectures and compression modes under the same input and generation
scale.
Specifically, we use the Baseline Large token workload as the common
reference, avoiding confounding effects from model-dependent generation
lengths.

For Multi-Ratio ConceptMoE, prefill and decode computation are scaled using
the compression ratios measured from evaluation telemetry, rather than
assuming the nominal CR2, CR4, or CR8 values.

The FLOPs reported in the main paper include full attention computation,
including QK$^\top$, attention--value multiplication, and sliding-window
attention.
For prefill, we follow an inference-serving LM-head convention and do not
apply the LM head at every prompt position; during decoding, the LM head is
computed normally for each newly generated token.

The reported FLOPs measure algorithmic computation under a standardized
workload and should not be interpreted as wall-clock latency, throughput, or
energy consumption.
Actual serving performance may additionally depend on kernel
implementations, memory access, communication, and batching strategies.

\subsection{Realized Compression Ratios}
\label{app:realized_cr}

CR2, CR4, and CR8 denote target average compression ratios rather than
fixed-length chunking rules.
Because chunk boundaries are content-adaptive, the realized compression ratio
may vary across workloads and independently trained checkpoints.
For a token sequence of length $N$, the effective concept-sequence length can
be approximated as
\[
N_{\mathrm{concept}}
\approx
\frac{N}{r_{\mathrm{realized}}}.
\]

For FLOPs accounting, we therefore use the measured prefill compression ratio
$r_{\mathrm{prefill}}$ and decode compression ratio $r_{\mathrm{decode}}$
for each checkpoint and evaluation workload.
The reported ratios are token-weighted averages over the terminal-turn
workloads defined in Appendix~\ref{app:terminal_flops}.
Table~\ref{tab:realized_cr} reports the values used in the CodeChat and CoSQL
FLOPs calculations.

\begin{table}[t]
\centering
\small
\caption{
\textbf{Workload-realized compression ratios.}
$r_{\mathrm{prefill}}$ and $r_{\mathrm{decode}}$ are the measured compression
ratios used for FLOPs accounting.
}
\label{tab:realized_cr}
\setlength{\tabcolsep}{4pt}
\begin{tabular}{llcc}
\toprule
Dataset & Model & $r_{\mathrm{prefill}}$ & $r_{\mathrm{decode}}$ \\
\midrule

\multirow{7}{*}{CodeChat}
& Multi-Ratio CR2   & 2.91 & 3.82 \\
& Multi-Ratio CR4   & 4.70 & 5.40 \\
& Multi-Ratio CR8   & 6.81 & 8.09 \\
& Multi-Ratio Mixed & 3.69 & 5.69 \\
& SingleCR2         & 1.54 & 1.60 \\
& SingleCR4         & 3.03 & 3.24 \\
& SingleCR8         & 7.67 & 12.27 \\
\midrule

\multirow{7}{*}{CoSQL}
& Multi-Ratio CR2   & 1.95 & 2.38 \\
& Multi-Ratio CR4   & 3.56 & 4.66 \\
& Multi-Ratio CR8   & 5.53 & 6.19 \\
& Multi-Ratio Mixed & 2.28 & 3.36 \\
& SingleCR2         & 1.88 & 1.31 \\
& SingleCR4         & 2.82 & 2.82 \\
& SingleCR8         & 3.62 & 3.32 \\
\bottomrule
\end{tabular}
\end{table}

These workload-realized ratios, rather than the nominal CR labels, are used
when computing the FLOPs reported in the main paper.
Consequently, checkpoints with the same nominal CR can have different
workload-realized FLOPs.

\subsection{Routed Independent-Model Switching}
\label{app:routed_flops}

Section~\ref{sec:model_switching} compares Multi-Ratio ConceptMoE with a
turn-wise routing baseline based on independent models.
Both methods follow the same compute-budget schedule, with
\[
\mathrm{CR2}\rightarrow\mathrm{Large},
\qquad
\mathrm{CR4}\rightarrow\mathrm{Medium},
\qquad
\mathrm{CR8}\rightarrow\mathrm{Small}.
\]

Both methods use the terminal-turn workload defined in
Appendix~\ref{app:terminal_flops} as their base computation.
Since CR2, CR4, and CR8 are operating modes of the same Multi-Ratio model,
changing the compression mode does not introduce cross-model state
incompatibility.
In contrast, Large, Medium, and Small are independent models with
model-specific history states.
For the routed baseline, we account for the history re-prefill required to
reconstruct the target model state after each model switch:
\[
F_{\mathrm{routed}}
=
F_{\mathrm{terminal}}
+
\sum_{j\in\mathcal S}
F_{\mathrm{re\text{-}prefill}}^{(j)},
\]
where $\mathcal S$ denotes the set of model-switch events.

These additional re-prefill operations are used only to reconstruct the KV
state of the target model, and therefore do not apply the LM head at every
re-prefilled position.
This keeps the same base FLOPs convention for both methods while accounting
only for the additional history-state reconstruction cost introduced by
independent-model switching.
This analysis isolates the additional algorithmic FLOPs associated with
state reconstruction under model switching.

\begin{table}[t]
\centering
\small
\caption{
\textbf{Switching workload for the routed independent-model baseline.}
Base FLOPs follow the same terminal-turn workload used throughout the
multi-turn evaluation, while re-prefill FLOPs account for history-state
reconstruction after model switches.
}
\label{tab:routed_flops_breakdown}
\begin{tabular}{lrrrrrr}
\toprule
Dataset
& \shortstack{Prompt\\Tokens}
& \shortstack{Output\\Tokens}
& Switches
& \shortstack{Re-prefilled\\Tokens}
& \shortstack{Base\\FLOPs}
& \shortstack{Re-prefill\\FLOPs}
\\
\midrule
CodeChat
& 923,286
& 59,910
& 1,925
& 2,882,580
& 1.5530
& 3.5365
\\

CoSQL
& 214,740
& 15,092
& 512
& 320,108
& 0.2830
& 0.3419
\\
\bottomrule
\end{tabular}
\end{table}

The resulting routed totals are 5.0895 PFLOPs on CodeChat and
0.6250 PFLOPs on CoSQL, compared with 1.0999 and 0.3319 PFLOPs for
Multi-Ratio Mixed, respectively.
This comparison quantifies the additional switching overhead avoided by
integrating multiple compute operating points within one shared model.

\section{Full General-Benchmark Results}
\label{app:general_results}

\subsection{Aggregate Results}
\label{app:aggregate_results}

Table~\ref{tab:full_general_results} reports the aggregate results of the
three fixed compute operating points on PTBench and SFTBench.
We compare the MoE baselines, independently trained SingleCR specialists,
and the fixed modes of Multi-Ratio ConceptMoE.

\begin{table}[t]
\centering
\small
\caption{
\textbf{Full general-benchmark results across compute budgets.}
For the MoE baselines, High/Medium/Low correspond to the operating points
matched to CR2/CR4/CR8, respectively.
The Avg. rows average over the three operating points.
}
\label{tab:full_general_results}
\begin{tabular}{llcc}
\toprule
Model & Operating Point & PTBench & SFTBench \\
\midrule

\multirow{4}{*}{MoE Baseline}
& High   & 31.7 & 19.0 \\
& Medium & 28.2 & 15.5 \\
& Low    & 25.4 & 13.1 \\
& Avg.   & 28.4 & 15.9 \\
\midrule

\multirow{4}{*}{SingleCR Specialist}
& CR2  & \textbf{33.2} & 21.1 \\
& CR4  & \textbf{30.0} & 18.8 \\
& CR8  & 25.8 & 16.4 \\
& Avg. & 29.7 & 18.8 \\
\midrule

\multirow{4}{*}{Multi-Ratio}
& CR2  & 32.1 & \textbf{21.5} \\
& CR4  & 29.3 & \textbf{19.1} \\
& CR8  & \textbf{27.6} & \textbf{17.2} \\
& Avg. & 29.7 & \textbf{19.3} \\
\bottomrule
\end{tabular}
\end{table}

\subsection{Category-Level Breakdown}
\label{app:category_breakdown}

\begin{table}[t]
\centering
\caption{
\textbf{Category-level PTBench results.}
All entries report accuracy (\%).
Overall is the weighted aggregate used in the main paper.
}
\label{tab:ptbench_category}

\small
\renewcommand{\arraystretch}{1.0}

\makebox[\textwidth][c]{%
\begin{tabular*}{1.0\textwidth}{
@{\extracolsep{\fill}}
lccccccc
@{}
}
\toprule
Model &
Comp. &
Reason. &
Math &
Code &
Spec. Know. &
Know. &
Overall \\
\midrule

MoE Large
& 41.5 & 25.6 & 12.1 & 46.9 & 26.3 & 26.6 & 31.7 \\
Multi-Ratio CR2
& 42.2 & 26.3 & 11.6 & 46.8 & 29.8 & 24.9 & 32.1 \\
SingleCR2
& 42.6 & 27.7 & 12.6 & 47.7 & 29.2 & 29.1 & 33.2 \\
\midrule

MoE Medium
& 36.5 & 20.8 & 9.3 & 42.5 & 26.1 & 25.2 & 28.2 \\
Multi-Ratio CR4
& 39.2 & 22.4 & 10.6 & 43.5 & 27.9 & 21.9 & 29.3 \\
SingleCR4
& 39.8 & 22.0 & 11.1 & 45.2 & 25.4 & 25.6 & 30.0 \\
\midrule

MoE Small
& 33.3 & 16.4 & 7.5 & 39.8 & 25.1 & 22.3 & 25.4 \\
Multi-Ratio CR8
& 35.8 & 19.5 & 9.3 & 41.1 & 27.6 & 23.9 & 27.6 \\
SingleCR8
& 33.2 & 17.9 & 8.8 & 42.6 & 25.3 & 19.4 & 25.8 \\

\bottomrule
\end{tabular*}%
}
\end{table}

\begin{table}[t]
\centering
\caption{
\textbf{Category-level SFTBench results.}
All entries report accuracy (\%).
Overall is the weighted aggregate used in the main paper.
}
\label{tab:sftbench_category}

\small
\renewcommand{\arraystretch}{1.0}

\makebox[\textwidth][c]{%
\begin{tabular*}{1.0\textwidth}{
@{\extracolsep{\fill}}
lcccccccc
@{}
}
\toprule
Model &
Comp. &
Reason. &
Math &
Code &
Inst. Follow. &
Know. &
Multi. &
Overall \\
\midrule

MoE Large
& 28.0 & 5.5 & 5.9 & 4.0 & 28.9 & 18.3 & 49.3 & 19.0 \\
Multi-Ratio CR2
& 30.4 & 8.2 & 7.3 & 8.1 & 33.6 & 15.7 & 56.3 & 21.5 \\
SingleCR2
& 29.8 & 8.5 & 7.0 & 4.2 & 35.9 & 18.5 & 53.0 & 21.1 \\
\midrule

MoE Medium
& 24.6 & 3.2 & 2.6 & 4.2 & 21.3 & 14.0 & 40.3 & 15.5 \\
Multi-Ratio CR4
& 28.0 & 5.5 & 4.4 & 5.3 & 33.6 & 15.2 & 48.0 & 19.1 \\
SingleCR4
& 27.8 & 7.1 & 4.6 & 5.2 & 30.1 & 14.2 & 49.0 & 18.8 \\
\midrule

MoE Small
& 21.7 & 1.9 & 2.0 & 2.1 & 19.9 & 11.5 & 32.0 & 13.1 \\
Multi-Ratio CR8
& 25.3 & 4.2 & 4.6 & 4.8 & 32.6 & 14.2 & 37.0 & 17.2 \\
SingleCR8
& 24.0 & 6.2 & 3.3 & 4.4 & 30.5 & 10.7 & 41.3 & 16.4 \\

\bottomrule
\end{tabular*}%
}
\end{table}

Tables~\ref{tab:ptbench_category} and~\ref{tab:sftbench_category} report
the category-level results underlying the aggregate PTBench and SFTBench
scores.

\section{Additional Cross-Ratio Training Ablations}
\label{app:cross_ratio_ablation}

\subsection{Per-Ratio Fixed-Mode Results}
\label{app:cross_ratio_per_ratio}

Table~\ref{tab:cross_ratio_per_ratio} expands the aggregate results reported
in the main paper into CR2, CR4, and CR8.
Increasing cross-ratio exposure has mixed effects at the PT stage, while the
fully cross-ratio configuration achieves the strongest SFTBench result at
all three fixed operating points.

\begin{table}[t]
\centering
\small
\caption{
\textbf{Per-ratio breakdown of cross-ratio training exposure.}
All entries report accuracy (\%).
}
\label{tab:cross_ratio_per_ratio}
\begin{tabular}{llccc}
\toprule
Evaluation
& Mode
& \shortstack{3 Single\\0 Cross}
& \shortstack{1 Single\\2 Cross}
& \shortstack{0 Single\\3 Cross} \\
\midrule

\multirow{3}{*}{PTBench}
& CR2 & 32.4 & \textbf{32.5} & 32.1 \\
& CR4 & \textbf{29.9} & 29.8 & 29.3 \\
& CR8 & \textbf{27.6} & \textbf{27.6} & \textbf{27.6} \\
\midrule

\multirow{3}{*}{SFTBench}
& CR2 & 21.2 & 20.7 & \textbf{21.5} \\
& CR4 & 18.7 & 18.4 & \textbf{19.1} \\
& CR8 & 16.6 & 17.1 & \textbf{17.2} \\

\bottomrule
\end{tabular}
\end{table}

\subsection{Mixed-Inference Compute}
\label{app:cross_ratio_compute}

Table~\ref{tab:cross_ratio_compute} reports the corresponding mixed-ratio
accuracy and workload-realized FLOPs.
The gains from fully cross-ratio training are not explained by increased
inference computation: on both tasks, the final configuration achieves the
highest accuracy with FLOPs comparable to or lower than the alternatives.

\begin{table}[t]
\centering
\small
\caption{
\textbf{Mixed-ratio inference under different cross-ratio training exposure.}
CodeChat reports Combined Joint accuracy and CoSQL reports SQLite execution
accuracy. FLOPs include attention-map computation.
}
\label{tab:cross_ratio_compute}
\begin{tabular}{lcc|cc}
\toprule
& \multicolumn{2}{c|}{CodeChat}
& \multicolumn{2}{c}{CoSQL} \\
Training Composition
& Acc. (\%) & FLOPs (PF)
& Acc. (\%) & FLOPs (PF) \\
\midrule

3 Single + 0 Cross
& 37.65 & 1.1719
& 43.00 & 0.3559 \\

1 Single + 2 Cross
& 37.57 & 1.1601
& 45.88 & 0.3395 \\

0 Single + 3 Cross
& \textbf{43.78} & \textbf{1.0999}
& \textbf{51.84} & \textbf{0.3319} \\

\bottomrule
\end{tabular}
\end{table}

\section{Full-Token Architecture Design}
\label{app:architecture_design}

\subsection{Motivation}

The encoder and decoder in ConceptMoE operate at full token resolution, so
their computation does not decrease as the compression ratio increases.
If these modules account for a substantial fraction of the total compute,
aggressive compression in the concept backbone yields only limited end-to-end
savings.
We therefore aim to reduce this compression-independent compute floor while
preserving sufficient token-level processing capacity.

Two natural design choices are to reduce the depth or the hidden width of the
full-token encoder and decoder.
Because full-scale pretraining is expensive, we use early-stage optimization
behavior to screen candidate architectures before committing to complete
training runs.

\subsection{Early-Stage Architecture Screening}

We compare three representative allocations under a constrained full-token
compute budget.
For clarity, we report relative configurations rather than absolute model
dimensions.
The depth of the final encoder/decoder design is normalized to $1\times$, and
the concept-backbone hidden width is normalized to $1\times$.

\begin{table}[!htb]
\centering
\small
\caption{
\textbf{Early-stage screening of full-token encoder/decoder designs.}
Alternative configurations with substantially worse early optimization were
terminated before full training.
}
\label{tab:architecture_screening}
\begin{tabular}{lccc}
\toprule
Configuration
& E/D Depth
& E/D Width
& Early Training \\
\midrule
Shallow-wide
& $0.5\times$
& $1.0\times$
& Worse convergence \\

Deep-narrow
& $2.0\times$
& $0.5\times$
& Worse convergence \\

\textbf{Balanced (ours)}
& $1.0\times$
& $\sim0.67\times$
& Stable \\
\bottomrule
\end{tabular}
\end{table}

Both aggressive depth reduction and aggressive width reduction led to
noticeably worse optimization during early training and were therefore
terminated early.
Our final architecture adopts the balanced configuration, retaining moderate
token-level depth while reducing hidden width.
This design lowers the full-token compute floor while maintaining stable
optimization, and is used throughout the main experiments.

\end{document}